\documentclass[sigconf]{acmart}
\usepackage{graphicx}
\usepackage{latexsym}
\usepackage{algorithm}
\usepackage{algorithmic}
\usepackage{graphicx} % DO NOT CHANGE THIS
\usepackage{hyperref}
\usepackage{amsmath}
\usepackage{amsthm}

\usepackage{enumitem}
\usepackage{amsfonts}
\usepackage{newfloat}
\usepackage{listings}
\usepackage{array}
\usepackage{threeparttable}
\usepackage{diagbox}
\usepackage{multirow}
\usepackage{longtable}
\usepackage{supertabular}
\usepackage{booktabs}
\usepackage{colortbl}
\usepackage{arydshln}
\usepackage[normalem]{ulem}
\usepackage{tabularray}

\AtBeginDocument{%
  \providecommand\BibTeX{{%
    \normalfont B\kern-0.5em{\scshape i\kern-0.25em b}\kern-0.8em\TeX}}}

\setcopyright{acmcopyright}
\copyrightyear{2018}
\acmYear{2018}
\acmDOI{XXXXXXX.XXXXXXX}

\acmConference[Conference acronym 'XX]{Make sure to enter the correct
  conference title from your rights confirmation emai}{June 03--05,
  2018}{Woodstock, NY}

\begin{document}

\title{OpenGDA: Graph Domain Adaptation Benchmark for Cross-network Learning}

\author{Boshen Shi}
\affiliation{%
  \institution{Data Intelligence System Research Center, Institute of Computing Technology, Chinese Academy of Sciences, University of Chinese Academy of Sciences}
  \city{Beijing}
  \country{China}}
\email{shiboshen19s@ict.ac.cn}

\author{Yongqing Wang}
\affiliation{%
  \institution{Data Intelligence System Research Center, Institute of Computing Technology}
  \city{Beijing}
  \country{China}}
\email{wyq@ict.ac.cn}

\author{Fangda Guo}
\affiliation{%
  \institution{Data Intelligence System Research Center, Institute of Computing Technology}
  \city{Beijing}
  \country{China}}
\email{guofangda@ict.ac.cn}

\author{Jiangli Shao}
\affiliation{%
  \institution{Data Intelligence System Research Center, Institute of Computing Technology, University of Chinese Academy of Sciences}
  \city{Beijing}
  \country{China}}
\email{shaojiangli19z@ict.ac.cn}

\author{Huawei Shen}
\affiliation{%
  \institution{Data Intelligence System Research Center, Institute of Computing Technology}
  \city{Beijing}
  \country{China}}
\email{shenhuawei@ict.ac.cn}

\author{Xueqi Cheng}
\affiliation{%
  \institution{Data Intelligence System Research Center, Institute of Computing Technology}
  \city{Beijing}
  \country{China}}
\email{cxq@ict.ac.cn}

\begin{abstract}
Graph domain adaptation models are widely adopted in cross-network learning tasks, with the aim of transferring labeling or structural knowledge. Currently, there mainly exist two limitations in evaluating graph domain adaptation models. On one side, they are primarily tested for the specific cross-network node classification task, leaving tasks at edge-level and graph-level largely under-explored. Moreover, they are primarily tested in limited scenarios, such as social networks or citation networks, lacking validation of model's capability in richer scenarios. As comprehensively assessing models could enhance model practicality in real-world applications, we propose a benchmark, known as \textit{OpenGDA}. It provides abundant pre-processed and unified datasets for different types of tasks (node, edge, graph). They originate from diverse scenarios, covering web information systems, urban systems and natural systems. Furthermore, it integrates state-of-the-art models with standardized and end-to-end pipelines. Overall, \textit{OpenGDA} provides a user-friendly, scalable and reproducible benchmark for evaluating graph domain adaptation models. The benchmark experiments highlight the challenges of applying GDA models to real-world applications with consistent good performance, and potentially provide insights to future research. As an emerging project, \textit{OpenGDA} will be regularly updated with new datasets and models. It could be accessed from https://github.com/Skyorca/OpenGDA.
\end{abstract}




\maketitle

\section{Introduction}
\label{intro}
% cross network learning task; gda; application.
% 点出评测的difficulty：数据集的准备。造成了目前任务和场景比较单一的问题。这不仅造成了算法间比较不充分，还对面向边和图级任务的研究造成了困难。
% 两个目标：1）任务和数据集 2）算法集成。用上一系列工程化的词。
Real-world graph data often faces the problem of limited labeling and sparse structures, which will degrade the performance of graph models~\cite{nolabel,tail}. To mitigate such problem and improve task performance, researchers establish cross-network learning tasks for leveraging relevant source graphs to transfer abundant labeling or structural knowledge to target graphs~\cite{fang2013transfer,cdne}. As source and target graphs may originate from correlated yet distinct domains, such as road networks from multiple regions, there exist both node feature distribution shift and graph structure shift between them~\cite{graden}. Therefore, graph domain adaptation(GDA) has been proposed to overcome distribution shifts and effectively transfer knowledge~\cite{acdne,adagcn,dane,udagcn,asn,specgda,baoligda}. Inspired by conventional domain adaptation methods~\cite{dan,dann,dsn}, GDA adapts such technique to graphs by taking the unique properties of graph-structured data into account. As an emerging area of research, GDA has the potential to boost graph learning tasks in various real-world applications\cite{userprop1,userprop2,physical1,physical2,bio1, bio2}.

Experimentally assessing GDA models is important to understand their competencies. However, previous studies mainly have two limitations in evaluation, that is, the limited type of task and the insufficient quantity of scenarios. On one side, although most GDA models develop general framework for cross-network learning, they are only tested for the specific cross-network node classification task, leaving tasks at edge-level and graph-level largely under-explored. Furthermore, they are primarily tested in limited scenarios, such as social networks or citation networks, lacking validation of model's ability in diverse scenarios. These limitations largely stem from the challenges associated with collecting suitable datasets for different types of tasks, which meet the requisite criteria for GDA. These requirements dictate that each dataset should 1) comprise a group of relevant graphs originated from similar yet different domains and 2) ensure the feature spaces and label spaces of graphs are consistent with each other~\cite{yqsurvey,zfzsurvey}. As GDA is crucial to tackle cross-network learning tasks, it is necessary to comprehensively verify model capability by testing them in diverse scenarios for different types of tasks.

\begin{figure}[h]
  \centering
  \includegraphics[width=\linewidth]{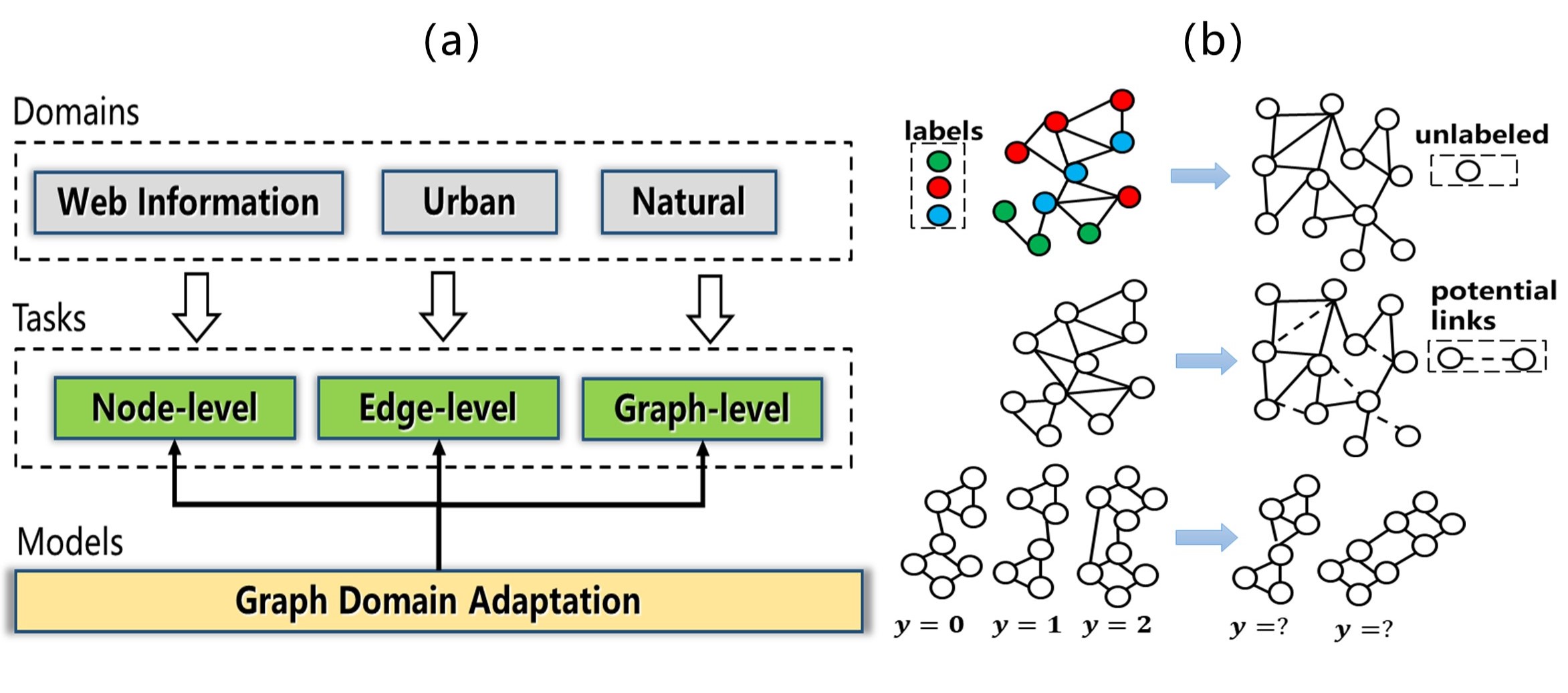}
  \caption{\textbf{(a)}: OpenGDA provides datasets that are diverse in scenarios and tasks, and integrates standardized GDA models for evaluation. \textbf{(b)}: Examples for cross-network learning tasks on node-level, edge-level and graph-level.}
  \label{figidea}
\end{figure}

To address such limitations, we propose a systematic graph domain adaptation benchmark in this work, known as \textit{OpenGDA}. As design principles, we strive to 1) provide abundant datasets from diverse scenarios with different cross-network learning tasks; and 2) integrate SOTA GDA models for encouraging a fair and comprehensive comparison. Specifically, \textit{OpenGDA} currently provides four node-level datasets, three edge-level datasets and two graph-level datasets, covering diverse scenarios including web information systems, urban systems and natural systems. Overall, \textit{OpenGDA} provides 70+ cross-network learning tasks. The details of datasets are summarized in Table~\ref{dataset}. Apart from tasks and datasets, \textit{OpenGDA} also pre-implements six SOTA GDA models~\cite{dane,adagcn,udagcn,asn,graden,specgda} and some baseline models, which are based on graph neural network (GNN). It is worth mentioning that we adopt PyTorch and PyTorch Geometric(PyG)~\cite{pyg} to standardize the overall pipeline for each model, including data interface, model architechture, and training/evaluation. Consequently, \textit{OpenGDA} is highly unified and customizable, supporting adding new datasets or models. The experimental results based on \textit{OpenGDA} highlight the challenges of applying GDA models to real-world applications with a steady good performance, and may provide insights to guide future research in this field. \textit{OpenGDA} will be regularly updated with new dataset and models. Generally, the contributions of this work is summarized as follows.
\begin{itemize}
    \item To the best of our knowledge, \textit{OpenGDA} is the first benchmark for evaluating GDA models comprehensively.
    \item \textit{OpenGDA} provides abundant datasets, aiming to assess the capability of GDA models to handle different cross-network learning tasks in diverse scenarios. These datasets are collected, pre-processed and standardized for the convenience of researchers.
    \item \textit{OpenGDA} integrates multiple SOTA GDA models which are standardized from each pipeline stage. Overall, the numerical results from benchmark experiments provide foundations for future research developments.
\end{itemize}

\begin{table*}
\centering
\caption{Details of tasks and currently-available datasets provided by OpenGDA. \textit{\#Domains} indicates the number of domains covered in the dataset, and \textit{\#Tasks} refers to the number of corresponding cross-network learning tasks built between domains. Dataset statistics, such as \textit{\#Feat}, illustrate the statistics for each domain, and we use '-' to connect minimum and maximum values when statistic varies across domains.}
\label{dataset}
\resizebox{\linewidth}{!}{%
\begin{tblr}{
  cells = {c},
  cell{2}{1} = {r=4}{},
  cell{2}{2} = {r=4}{},
  cell{6}{1} = {r=3}{},
  cell{6}{2} = {r=2}{},
  cell{9}{1} = {r=2}{},
  cell{9}{2} = {r=2}{},
  hline{1-2,6,9} = {-}{},
  hline{11} = {-}{0.08em},
}
Task Level  & Task Type            & Dataset              & \#Domains & \#Tasks & \#Feat & \#Label & \#Nodes      & \#Edges        & \#Graphs  \\
Node-level  & node classification  & Citation1~\cite{acdne}          & 3         & 6       & 6775   & 5       & 5484 - 9360  & 8130-15602     & 1         \\
            &                      & Twitch~\cite{twitch}              & 6         & 30      & 3170   & 1       & 1912-9498    & 31299 - 112667 & 1         \\
            &                      & Blog~\cite{acdne}                & 2         & 2       & 8189   & 6       & 2300-2896    & 33471-53836    & 1         \\
            &                      & Airport~\cite{graden}             & 3         & 6       & 8      & 4       & 131-1190     & 1038-13599     & 1         \\
Edge-level  & link prediction      & Amazon Review~\cite{graden}       & 4         & 8       & 5000   & 1       & 8568 - 95248 & 51190-353942   & 1         \\
            &                      & Citation2~\cite{udagcn}           & 2         & 2       & 7537   & 1       & 5578-7410    & 7341-11135     & 1         \\
            & link classification  & PPI~\cite{specgda}                 & 5         & 20      & 256      & 2       & 4286-8369    & 55668-207461   & 1         \\
Graph-level & graph classification & IMDB-REDDIT~\cite{tudataset}         & 2         & 2       & 136    & 1       & 19773-859254 & 96531-995508   & 1000-2000 \\
            &                      & LetterHigh-LetterLow~\cite{tudataset} & 2         & 2       & 2      & 15      & 10507-10522  & 14092-20250    & 2250      
\end{tblr}
}
\end{table*}

\section{Related Work}
Graph domain adaptation utilizes both source and target graphs for training, and tests model mainly on target graphs. Generally, both methods adopt deep graph models, such as GNN, to integrate node feature distribution shift and graph structure shift together as node embedding distribution shit. It could be classified into discrepancy-based methods~\cite{acdne,udagcn,adagcn,graden,specgda,baoligda} and disentangle-based methods~\cite{asn,gengda}. For discrepancy-based methods, they compute such distribution shit via discrepancy measurement and gradually reduce it with supervision loss. Consequently, the knowledge from source graphs could be transferred when both domain discrepancy and supervision loss are minimized. Many discrepancy-based methods tend to use conventional discrepancy measurement like Wasserstein distance or Jensen-Shannon distance, while GRADE~\cite{graden} and SpecReg~\cite{specgda} further improves discrepancy measurement by taking graph properties into account. For disentangle-based methods, they generally disentangle node embeddings into domain-invariant and domain-relevant parts. Subsequently, they transfer knowledge with domain-invariant embeddings by minimizing supervision loss. 

Currently, many benchmarks have been developed for graph learning tasks. Three main categories of graph learning benchmarks include 1) \textbf{Benchmarks for general graph machine learning}, such as Benchmarking-gnns~\cite{gnnbench} and Open Graph Benchmark~\cite{ogb}, 2) \textbf{Benchmarks for out-of-distribution on graphs}, such as GOOD~\cite{good} and DrugOOD~\cite{drugood}, and 3) \textbf{Benchmarks for self-supervised learning on graphs}, such as DIG~\cite{dig}. As these benchmarks split train and test data from the same domain, they do not satisfy the settings of cross-network learning and GDA. Moreover, it's not trivial to directly establish tasks or datasets for GDA models from these benchmarks, because datasets used by GDA models have restrictions discussed in Section~\ref{intro}. Therefore, it's necessary to establish a benchmark for evaluating GDA models for cross-network learning tasks. 

\section{Benchmark Design}

Generally, designing \textit{OpenGDA} involves two key stages. The first is preparing abundant datasets from diverse scenarios, which satisfy the settings for GDA. Initially, we collect popular datasets in previous studies by task type. As most of raw datasets have distinct properties and inconsistent formats, we preprocess and unify these datasets, providing user-friendly graph objects. They are compatible with PyTorch and PyG. In addition, we identify potential datasets for evaluating GDA models from other research areas, such as graph out-of-distribution. As they may not entirely meet GDA settings, we carry out additional pre-processing concerning node features and labels. The format of these datasets is also unified with graph object.

\begin{figure}[h]
  \centering
  \includegraphics[width=\linewidth]{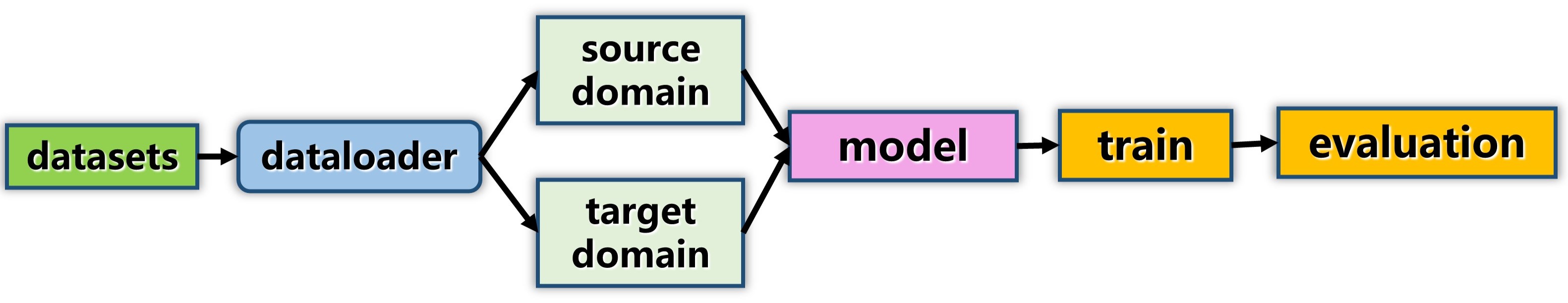}
  \caption{The standardized pipeline of OpenGDA. }
  \label{figpipeline}
\end{figure}

After the preparation of datasets, we standardize the overall pipeline for each model, including data interface, model architechture and training/evaluation. Figure~\ref{figpipeline} illustrates the standardized pipeline. \textbf{Data interface}: As discussed above, datasets are provided with unified graph object, each consisting of node feature matrix, adjacency matrix, label matrix and other related properties. They are loaded via a dataloader, which is shared across datasets and tasks. \textbf{Model architechture}: We establish models upon standard PyTorch modules and PyG GNN layers. In addition, they follow the same forward propagation process, in which they take both source and target domains as input and then compute two primary losses (i.e., supervision loss and domain discrepancy loss). \textbf{Training/Evaluation}: We employ standard PyTorch backward propagation for training, and define a suite of universal evaluation functions for diverse tasks. Consequently, \textit{OpenGDA} is user-friendly and supports either conducting experiments on pre-implemented elements or integrating new datasets and models. 

\begin{figure}[h]
  \centering
  \includegraphics[width=\linewidth]{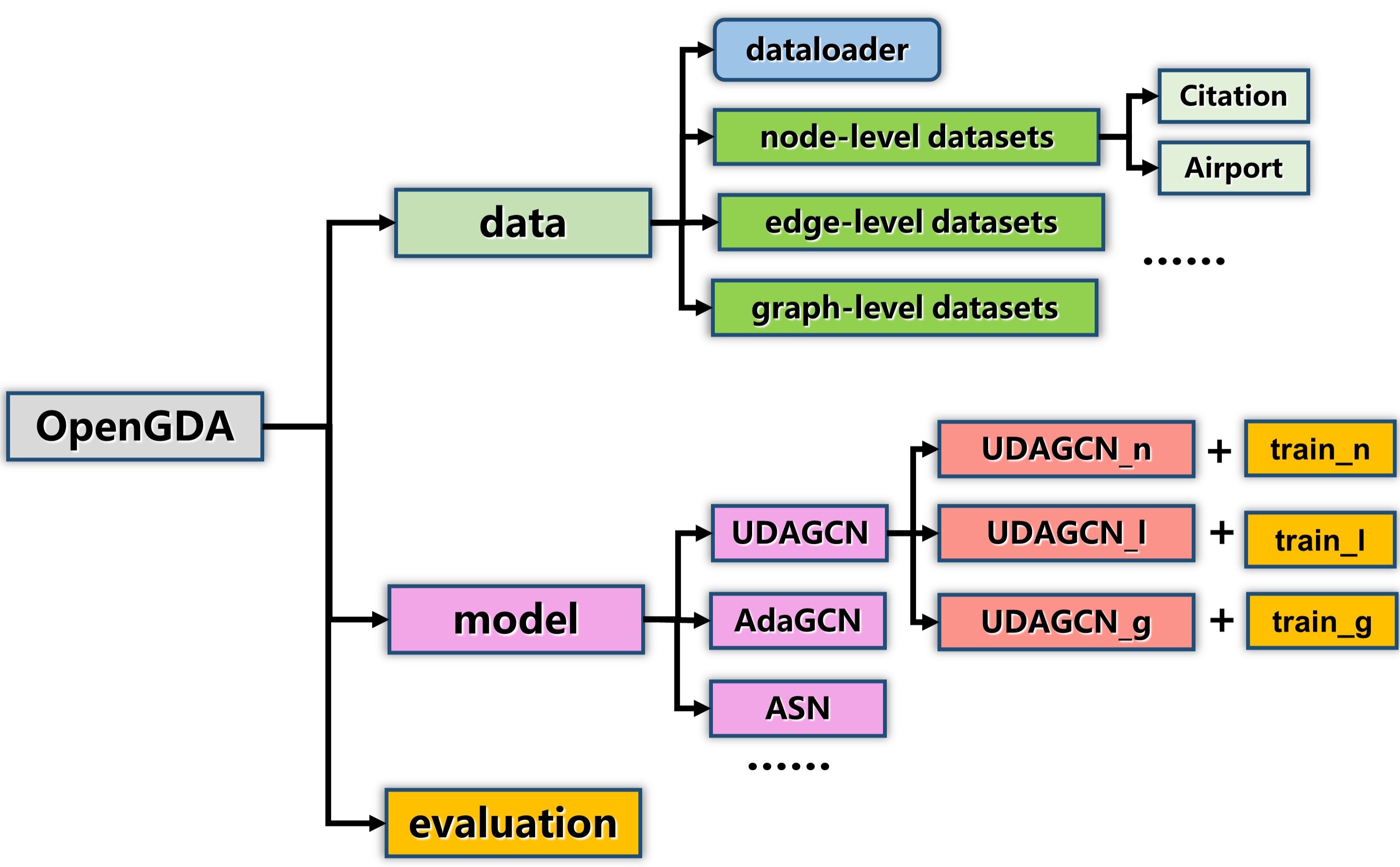}
  \caption{The framework of OpenGDA package. The \textit{data} module includes a dataloader and three folders to store pre-processed datasets based on task type. The \textit{model} module includes GDA models, each having three variants and the corresponding training manuscript. The \textit{evaluation} module includes a suite of metric functions.}
  \label{figframework}
\end{figure}

The \textit{OpenGDA} package is designed to make the pipeline of Figure~\ref{figpipeline} easily accessible to researchers, and the package framework is demonstrated in Figure~\ref{figframework}. Firstly, the pre-processed datasets are organized in different paths depending on their task categories, yet they share a common data-loader. Besides, as different types of tasks may require different forward propagation or training process, each model \textit{M} is implemented via three variants accordingly (i.e., \textit{M\_n}, \textit{M\_l} and \textit{M\_g} ). For example, a unique property of edge-level task is that some datasets contain bipartite graphs, such as Amazon Review dataset. Therefore, the model should separately consider two categories of nodes for domain adaptation, leading to a different forward propagation process compared to node-level and graph-level tasks. In general, the package is implemented with comprehensive file structures and code structures. It has a user-friendly workflow and could scale well with the addition of new models and datasets.

\section{Experimental Studies}
\begin{table*}
\centering
\caption{Node classification accuracy on Airport and Citation1 dataset. U, B and E are short for USA, Brazil and Europe in Airport dataset, while A, D and C are short for Acmv9, Dblpv7 and Citationv1 in Citation1 dataset. }
\label{node1}
%\resizebox{\linewidth}{!}{%
\begin{tabular}{l|llllll|llllll} 
\hline
       & \multicolumn{6}{c|}{Airport}                                                                                                                                                                                    & \multicolumn{6}{c}{Citation1}                                                                                                                                                                                    \\ 
\hline
       & U$\rightarrow$B & U$\rightarrow$E & B$\rightarrow$U & B$\rightarrow$E & E$\rightarrow$U & E$\rightarrow$B & A$\rightarrow$D & D$\rightarrow$A & A$\rightarrow$C & C$\rightarrow$A & C$\rightarrow$D & D$\rightarrow$C  \\ 
\hline
GCN    & 0.427                            & 0.436                            & 0.454                            & 0.481                            & 0.458                            & 0.465                            & 0.623                            & 0.578                            & 0.675                            & 0.635                            & 0.666                            & 0.654                             \\
UDAGCN & \textbf{0.607}                   & \textbf{0.488}                   & 0.497                            & \textbf{0.510}                   & 0.434                            & 0.477                            & 0.684                            & 0.623                            & 0.728                            & 0.663                            & 0.712                            & 0.645                             \\
AdaGCN & 0.466                            & 0.434                            & \textbf{0.501}                   & 0.486                            & 0.456                            & 0.561                            & 0.687                            & 0.663                            & 0.701                            & 0.643                            & 0.709                            & 0.702                             \\
ASN    & 0.519                            & 0.469                            & 0.498                            & 0.494                            & \textbf{0.466}                   & \textbf{0.595}                   & \textbf{0.709}                   & \textbf{0.703}                   & 0.732                            & 0.658                            & \textbf{0.732}                   & \textbf{0.734}                    \\
GRADE  & 0.550                            & 0.457                            & 0.497                            & 0.506                            & 0.463                            & 0.588                            & 0.701                            & 0.660                            & \textbf{0.736}                   & \textbf{0.687}                   & 0.722                            & 0.687                             \\
\hline
\end{tabular}
%}
\label{node1}
\end{table*}

\begin{table*}
\centering
\caption{Link prediction results on Amazon Review-nonoverlapping dataset, where users are disjoint between domains. }
\label{edge1}
\resizebox{\linewidth}{!}{%
\begin{tabular}{c|ccc|ccc|ccc|ccc} 
\hline
       & \multicolumn{3}{c|}{cd$\rightarrow$music}        & \multicolumn{3}{c|}{~music$\rightarrow$cd}       & \multicolumn{3}{c|}{~book$\rightarrow$movie}     & \multicolumn{3}{c}{~movie$\rightarrow$book}       \\ 
\hline
       & Hits@10        & MRR@10         & NDCG@10        & Hits@10        & MRR@10         & NDCG@10        & Hits@10        & MRR@10         & NDCG@10        & Hits@10        & MRR@10         & NDCG@10         \\ 
\hline
GCN    & 0.158          & 0.052          & 0.076          & 0.273          & 0.119          & 0.153          & 0.339          & 0.150          & 0.192          & 0.094          & 0.044          & 0.056           \\
UDAGCN & 0.376          & \textbf{0.155} & 0.206          & 0.255          & 0.102          & 0.136          & 0.369          & 0.149          & 0.197          & 0.194 & 0.103 & 0.124  \\
AdaGCN & 0.274          & 0.137          & 0.170          & 0.280          & 0.125          & 0.159          & 0.337          & 0.131          & 0.177          & 0.184          & 0.097          & 0.118           \\
ASN    & \textbf{0.380} & \textbf{0.155} & \textbf{0.207} & \textbf{0.285} & \textbf{0.131} & \textbf{0.165} & 0.348	&0.157	& 0.199	& \textbf{0.539} & \textbf{0.136} & \textbf{0.227} \\
GRADE  & 0.207          & 0.081          & 0.110          & 0.251          & 0.108          & 0.139          & \textbf{0.406} & \textbf{0.204} & \textbf{0.247} & 0.078          & 0.026          & 0.037           \\
\hline
\end{tabular}
}
\label{edge1}
\end{table*}
We conduct experiments to comprehensively evaluate GDA models based on \textit{OpenGDA}. We not only expect that the experiments could comprehensively validate GDA models from the aspect of diverse tasks and datasets, but also hope that the experimental results could shed light on the follow-up work in this field. 

Without loss of generality, we select four representative GDA models with naive GCN~\cite{gcn} baseline for evaluation in this work. Among these GDA models, UDAGCN and AdaGCN are widely adopted as baselines for their effectiveness and efficiency. GRADE represents the methods which further improve discrepancy measurement with graph properties, while ASN represents disentangle-based methods. Subsequently, we select tasks from node-level, edge-level and graph-level, each containing two datasets covering diverse scenarios. For node-level and edge-level tasks, one domain (source or target) comprises one graph, while it contains a group of graphs in graph-level tasks. We build cross-network learning tasks with one source domain and one target domain. In particular, we employ global mean pooling to compute graph embedding in graph-level tasks. During training, labeling information in target domain is not available in node-level and graph-level tasks. For evaluation, we adopt hit ratio (Hits@k), mean reciprocal rank (MRR@k), and normalized discounted cumulative gain (NDCG@k) where k equals 10 in edge-level tasks, and classification accuracy for other tasks. All models follow the same hyper-parameter settings. The corresponding experimental results are shown in Table~\ref{node1}-\ref{graph1}.

Overall, we observe two forms of inconsistency from numerical results. 1) \textbf{Scenario-inconsistency}, which indicates the model may fail to consistently perform well across datasets from various scenarios in specific tasks. For example, in edge-level tasks, UDAGCN tend to perform worse than ASN on Amazon Review dataset, but the situation is reversed on Citation2 dataset. 2) \textbf{Task-inconsistency}, which means model may fail to consistently perform well across different types of tasks. For example, AdaGCN performs averagely in egde-level tasks, but its performance improves significantly in node-level and graph-level tasks. These inconsistencies make it difficult to predict the performance of GDA models in real-world applications. Moreover, GDA models only slightly outperform GCN baseline in some tasks and scenarios. Such observations underscore the necessity for further enhancement of the practical applicability of GDA models. In addition to establishing benchmarks for comprehensive evaluation, such as \textit{OpenGDA}, more theoretical studies should be conducted for fully understanding and utilizing intrinsic graph structural properties and domain adaptation mechanism. 

\begin{table}
\centering
\caption{Link Prediction results on Citation2 dataset}
\label{edge2}
\resizebox{\linewidth}{!}{%
\begin{tabular}{l|lll|lll} 
\hline
       & \multicolumn{3}{c|}{acm$\rightarrow$dblp}                   & \multicolumn{3}{c}{dblp$\rightarrow$acm}                      \\ 
\hline
       & Hits@10        & MRR@10         & NDCG@10        & Hits@10        & MRR@10         & NDCG@10         \\ 
\hline
GCN    & 0.818          & 0.661          & 0.699          & 0.184          & 0.073          & 0.098           \\
UDAGCN & \textbf{0.853} & \textbf{0.675} & \textbf{0.716} & \textbf{0.258} & \textbf{0.108} & \textbf{0.141}  \\
AdaGCN & 0.533          & 0.401          & 0.432          & 0.145          & 0.045          & 0.068           \\
ASN    & 0.849          & 0.672          & 0.712          & 0.245          & 0.103          & 0.134           \\
GRADE  & 0.823          & 0.665          & 0.703          & 0.185          & 0.070          & 0.096           \\
\hline
\end{tabular}
}
\end{table}

\begin{table}
\centering
\caption{Graph classification accuracy on IMDB-REDDIT dataset. I, R, L and H are short for domain names.}
\label{graph1}
%\resizebox{\linewidth}{!}{%
\begin{tabular}{c|ccccc} 
\hline
    & GCN   & UDAGCN & AdaGCN & ASN            & GRADE  \\ 
\hline
I$\rightarrow$R & 0.537 & 0.564  & 0.568  & \textbf{0.589} & 0.565  \\
R$\rightarrow$I & 0.519 & 0.526  & 0.503  & \textbf{0.552} & 0.524  \\ 
\hline
L$\rightarrow$H & \textbf{0.121}     & 0.106      & 0.107      & 0.097              & 0.089      \\
H$\rightarrow$L & 0.092     & 0.093      & \textbf{0.136}      & 0.124              & 0.132      \\
\hline
\end{tabular}
%}
\end{table}

\section{Conclusion}
To enable a comprehensive evaluation of graph domain adaptation models for cross-network learning tasks, we develop graph domain adaptation benchmark, \textit{OpenGDA}, for validating model performance in diverse scenarios and different types of tasks, and encouraging a fair comparison across SOTA models. Through specially processed datasets and standardized pipelines, \textit{OpenGDA} offers user-friendly access to researchers for conducting experiments with pre-implemented elements and introducing additional datasets or models. The experimental results from \textit{OpenGDA} highlight the difficulties for achieving consistent good performance for existing GDA models. Altogether, \textit{OpenGDA} presents fruitful opportunities for future research. It's worth mentioning \textit{OpenGDA} is a growing project. We plan to integrate more methods as GDA is an emerging line of research, and include datasets and tasks of a larger quantity and variety. We also expect to keep improving package scalability and enhancing user experience.

\bibliographystyle{ACM-Reference-Format}
\bibliography{ref}
\end{document}